\documentclass[10pt,twocolumn,letterpaper]{article}

\usepackage{cvpr}              

\usepackage{algorithm}
\usepackage{algorithmic}
\usepackage{multirow}
\usepackage{tabularx}
\usepackage{amsfonts,bm}
\usepackage{threeparttable}
\usepackage{upgreek}
\usepackage{pifont}%
\usepackage{bbding}
\usepackage[normalem]{ulem}
\usepackage{makecell}
\usepackage{color}
\usepackage{xcolor}
\usepackage{colortbl}
\usepackage{wrapfig}
\usepackage{soul}
\usepackage{wasysym}
\usepackage{xspace}
\usepackage{subcaption}
\usepackage[first=0,last=9]{lcg}
\definecolor{Gray}{gray}{0.85}
\newcommand{\tabincell}[2]{\begin{tabular}{@{}#1@{}}#2\end{tabular}}  
\newcolumntype{x}[1]{>{\centering\arraybackslash}p{#1pt}}
\newcolumntype{y}[1]{>{\raggedright\arraybackslash}p{#1pt}}
\newcolumntype{z}[1]{>{\raggedleft\arraybackslash}p{#1pt}}

\definecolor{cvprblue}{rgb}{0.21,0.49,0.74}
\usepackage[pagebackref,breaklinks,colorlinks,citecolor=cvprblue]{hyperref}

\def\paperID{xxxx} 
\def\confName{ICCV}
\def\confYear{2025}

\title{YOLOv12: Attention-Centric Real-Time Object Detectors}

\author{Yunjie Tian\\
\textit{University at Buffalo}\\
{\tt\small \textit{yunjieti@buffalo.edu}}
\and
Qixiang Ye\\
\textit{University of Chinese Academy of Sciences}\\
{\tt\small \textit{qxye@ucas.ac.cn}}
\and
David Doermann\\
\textit{University at Buffalo}\\
{\tt\small \textit{doermann@buffalo.edu}}\\
\and
{\url{github.com/sunsmarterjie/yolov12}}\\
{\tt\small -- Technical Report --}
}

\begin{document}

\twocolumn[{
\renewcommand\twocolumn[1][]{#1}
\maketitle
\begin{center}
\centering
\includegraphics[width=.95\linewidth]{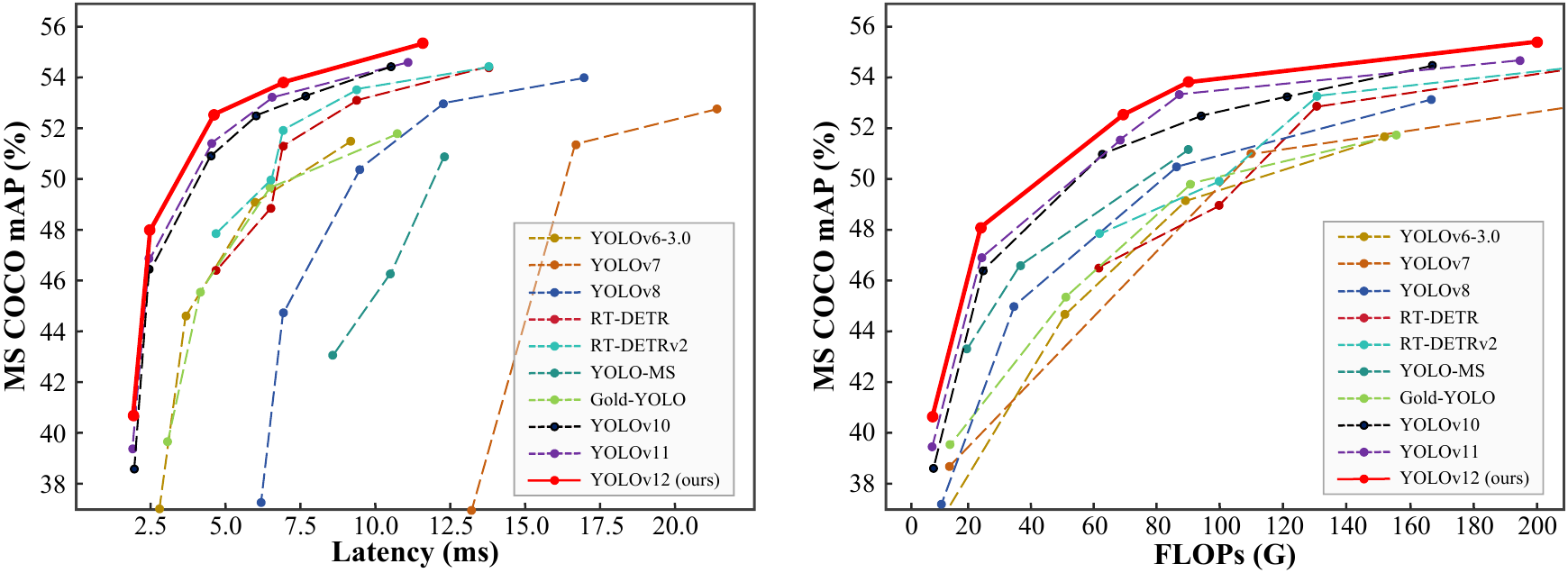}
\vspace{-2pt}
\captionof{figure}{\textbf{Comparisons with other popular methods in terms of latency-accuracy (left) and FLOPs-accuracy (right) trade-offs.}}
\label{fig:intro}
\end{center}
}]

\newcommand{\tick}{\textcolor{green}{\CheckmarkBold}\xspace}
\newcommand{\cross}{\textcolor{red}{\XSolidBrush}\xspace}

\begin{abstract}
Enhancing the network architecture of the YOLO framework has been crucial for a long time, but has focused on CNN-based improvements despite the proven superiority of attention mechanisms in modeling capabilities.
This is because attention-based models cannot match the speed of CNN-based models.
This paper proposes an attention-centric YOLO framework, namely YOLOv12, that matches the speed of previous CNN-based ones while harnessing the performance benefits of attention mechanisms.

YOLOv12 surpasses all popular real-time object detectors in accuracy with competitive speed. For example, YOLOv12-N achieves $40.6\%$ mAP with an inference latency of $1.64$ ms on a T4 GPU, outperforming advanced YOLOv10-N / YOLOv11-N by $2.1\%/1.2\%$ mAP with a comparable speed. This advantage extends to other model scales. 
YOLOv12 also surpasses end-to-end real-time detectors that improve DETR, such as RT-DETR / RT-DETRv2: YOLOv12-S beats RT-DETR-R18 / RT-DETRv2-R18 while running $42\%$ faster, using only $36\%$ of the computation and $45\%$ of the parameters.
More comparisons are shown in Figure~\ref{fig:intro}.
\end{abstract}

\section{Introduction}
\label{sec:intro}

Real-time object detection has consistently attracted significant attention due to its low-latency characteristics, which provide substantial practicality~\cite{yolov8, jocher2024yolov11, bogdoll2022anomaly, dos2019mobile}.
Among them, the YOLO series~\cite{redmon2016yolov1, redmon2017yolo9000, redmon2018yolov3, bochkovskiy2020yolov4, jocher2020yolov5, li2023yolov6, wang2023yolov7, yolov8, wang2024yolov9, wang2024yolov10, jocher2024yolov11} has effectively established an optimal balance between latency and accuracy, thus dominating the field.
Although improvements in YOLO have focused on areas such as loss functions~\cite{chen2020ap_loss, zhou2019iou_loss, oksuz2021rank_loss, oksuz2020ranking_loss, zheng2020distance_loss, li2020generalized_focal_loss, rezatofighi2019generalized_loss}, label assignment~\cite{ge2021ota_label, li2022dual_label, wang2021end_label, feng2021tood_label, zhu2020autoassign_label}, network architecture design has remained a critical research priority~\cite{li2023yolov6, wang2023yolov7, yolov8, wang2024yolov9, jocher2024yolov11}. 
Although attention-centric vision transformer (ViT) architectures have been proven to possess stronger modeling capabilities, even in small models~\cite{he2022masked, fang2024eva, fang2024eva02, tian2024fast}, most architectural designs continue to focus primarily on CNNs.

The primary reason for this situation lies in the inefficiency of the attention mechanism, which comes from two main factors: quadratic computational complexity and inefficient memory access operations of the attention mechanism (the latter being the main issue addressed by FlashAttention~\cite{dao2022flashattention, dao2023flashattentionv2}).
As a result, under a similar computational budget, CNN-based architectures outperform attention-based ones by a factor of $\sim3\times$~\cite{liu2024vmamba}, which significantly limits the adoption of attention mechanisms in YOLO systems where high inference speed is critical.

This paper aims to address these challenges and further builds an attention-centric YOLO framework, namely YOLOv12.
We introduce three key improvements. 
\textbf{First}, we propose a simple yet efficient area attention module (A2), which maintains a large receptive field while reducing the computational complexity of attention in a very simple way, thus enhancing speed.
\textbf{Second}, we introduce the residual efficient layer aggregation networks (R-ELAN) to address the optimization challenges introduced by attention (primarily large-scale models). R-ELAN introduces two improvements based on the original ELAN~\cite{wang2023yolov7}: \textbf{(i)} a block-level residual design with scaling techniques and \textbf{(ii)} a redesigned feature aggregation method.
\textbf{Third}, we make some architectural improvements beyond the vanilla attention to fit the YOLO system. We upgrade traditional attention-centric architectures including: introducing FlashAttention to conquer the memory access issue of attention, removing designs such as positional encoding to make the model fast and clean, adjusting the MLP ratio from $4$ to $1.2$ to balance the computation between attention and feed forward network for better performance, reducing the depth of stacked blocks to facilitate optimization, and making use of convolution operators as much as possible to leverage their computational efficiency.

Based on the designs outlined above, we develop a new family of real-time detectors with $5$ model scales: YOLOv12-N, S, M, L, and X. We perform extensive experiments on standard object detection benchmarks following YOLOv11~\cite{jocher2024yolov11} without any additional tricks, demonstrating that YOLOv12 provides significant improvements over previous popular models in terms of latency-accuracy and FLOPs-accuracy trade-offs across these scales, as illustrated in Figure~\ref{fig:intro}.
For example, YOLOv12-N achieves $40.6\%$ mAP, outperforming YOLOv10-N~\cite{wang2024yolov10} by $2.1\%$ mAP while maintaining a faster inference speed, and YOLOv11-N~\cite{jocher2024yolov11} by $1.2\%$ mAP with a comparable speed. This advantage remains consistent across other scale models. 
Compared to RT-DETR-R18~\cite{zhao2024rtdetrs} / RT-DETRv2-R18~\cite{lv2024rt_detrv2}, YOLOv12-S is $1.5\%/0.1\%$ mAP better, while reports $42\%/42\%$ faster latency speed, requiring only $36\%/36\%$ of their computations and $45\%/45\%$ of their parameters.

In summary, the contributions of YOLOv12 are two-fold: 1) it establishes an attention-centric, simple yet efficient YOLO framework that, through methodological innovation and architectural improvements, breaks the dominance of CNN models in YOLO series. 2) without relying on additional techniques such as pretraining, YOLOv12 achieves state-of-the-art results with fast inference speed and higher detection accuracy, demonstrating its potential.

\section{Related Work}
\label{sec:related}

\noindent\textbf{Real-time Object Detectors.} 
Real-time object detectors have consistently attracted the community's attention due to their significant practical value.
The YOLO series~\cite{redmon2016yolov1, redmon2017yolo9000, redmon2018yolov3, bochkovskiy2020yolov4, jocher2020yolov5, li2023yolov6, wang2023yolov7, wang2024gold, chen2023yolo, yolov8, wang2024yolov9, wang2024yolov10, jocher2024yolov11} has emerged as a leading framework for real-time object detection.
The early YOLO systems~\cite{redmon2016yolov1, redmon2017yolo9000, redmon2018yolov3} establish the framework for the YOLO series, primarily from a model design perspective. 
YOLOv4~\cite{bochkovskiy2020yolov4} and YOLOv5~\cite{jocher2020yolov5} add CSPNet~\cite{wang2020cspnet}, data augmentation, and multiple feature scales to the framework. YOLOv6~\cite{li2023yolov6} further advance these with BiC and SimCSPSPPF modules for the backbone and neck, alongside anchor-aided training. 
YOLOv7~\cite{wang2023yolov7} introduce E-ELAN~\cite{wang2022designing_elan} (efficient layer aggregation networks) for improved gradient flow and various bag-of-freebies, while YOLOv8~\cite{yolov8} integrate a efficient C2f block for enhanced feature extraction.
In recent iterations, YOLOv9~\cite{wang2024yolov9} introduce GELAN for architecture optimization and PGI for training improvement, while YOLOv10~\cite{wang2024yolov10} apply NMS-free training with dual assignments for efficiency gains. YOLOv11~\cite{jocher2024yolov11} further reduces latency and increases accuracy by adopting the C3K2 module (a specification of GELAN~\cite{wang2024yolov9}) and lightweight depthwise separable convolution in the detection head. 
Recently, an end-to-end object detection method, namely RT-DETR~\cite{zhao2024rtdetrs}, improved traditional end-to-end detectors~\cite{carion2020detr, liu2022dab_detr, zhu2020deformable_detr, meng2021conditional_detr, li2022dn_detr} to meet real-time requirements by designing an efficient encoder and an uncertainty-minimal query selection mechanism. RT-DETRv2~\cite{lv2024rt_detrv2} further enhances it with bag-of-freebies.
Unlike previous YOLO series, this study aims to build a YOLO framework centered around attention to leverage the superiority of the attention mechanism.

\begin{figure*}[t]
\centering
\includegraphics[width=0.99\linewidth]{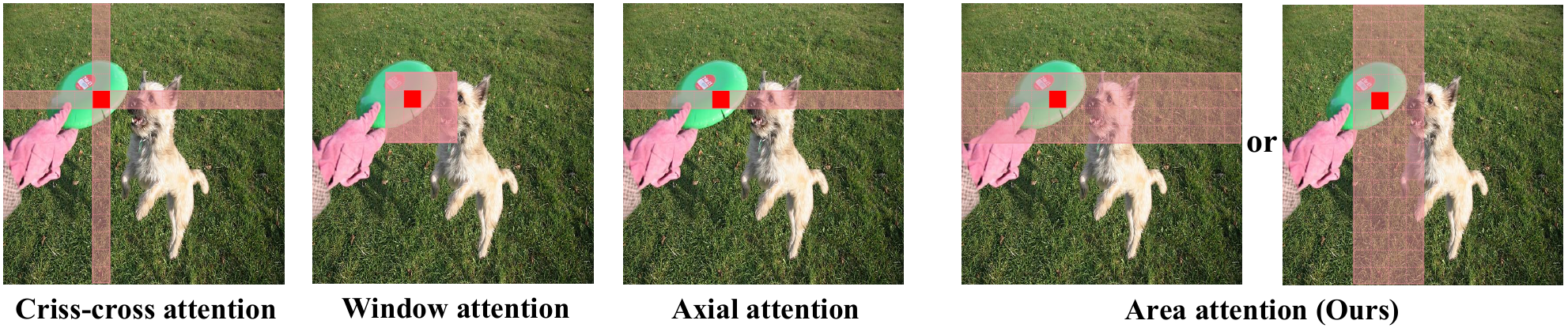}
\caption{\textbf{Comparison of the representative local attention mechanisms with our area attention.}  Area Attention adopts the most straightforward equal partitioning way to divide the feature map into $l$ areas vertically or horizontally. (default is 4). This avoids complex operations while ensuring a large receptive field, resulting in high efficiency.}
\label{fig:split}
\end{figure*}

\noindent\textbf{Efficient Vision Transformers.} 
Reducing computational costs from global self-attention is crucial to effectively applying vision transformers in downstream tasks. PVT~\cite{pvt} addresses this using multi-resolution stages and downsampling features. 
Swin Transformer~\cite{liu2021swin} limits self-attention to local windows and adjusts the window partitioning style to connect non-overlapping windows, balancing communication needs with memory and computation demands.
Other methods, such as axial self-attention~\cite{ho2019axial} and criss-cross attention~\cite{huang2019ccnet}, calculate attention within horizontal and vertical windows. CSWin transformer~\cite{dong2022cswin} builds on this by introducing cross-shaped window self-attention, computing attention along horizontal and vertical stripes in parallel.
In addition, local-global relations are established in works such as~\cite{chu2021twins, yu2021glance}, improving efficiency by reducing reliance on global self-attention.
Fast-iTPN~\cite{tian2024fast} improves downstream task inference speed with token migration and token gathering mechanisms.
Some approaches~\cite{shen2021linear_attn, wang2020linformer_attn, katharopoulos2020transformers_attn, xie2025sana} use linear attention to decrease the complexity of the attention. Although Mamba-based vision models~\cite{zhu2024vision_mamba, liu2024vmamba} aim for linear complexity, they still fall short of real-time speeds~\cite{liu2024vmamba}.
FlashAttention~\cite{dao2022flashattention, dao2023flashattentionv2} identifies high bandwidth memory bottlenecks that lead to inefficient attention computation and addresses them through I/O optimization, reducing memory access to enhance computational efficiency.
In this study, we discard complex designs and propose a simple area attention mechanism to reduce the complexity of attention. Additionally, we employ FlashAttention to overcome inherent memory accessing problems of the attention mechanism~\cite{dao2022flashattention, dao2023flashattentionv2}.

\begin{figure*}[!htb]
\centering
\includegraphics[width=0.99\linewidth]{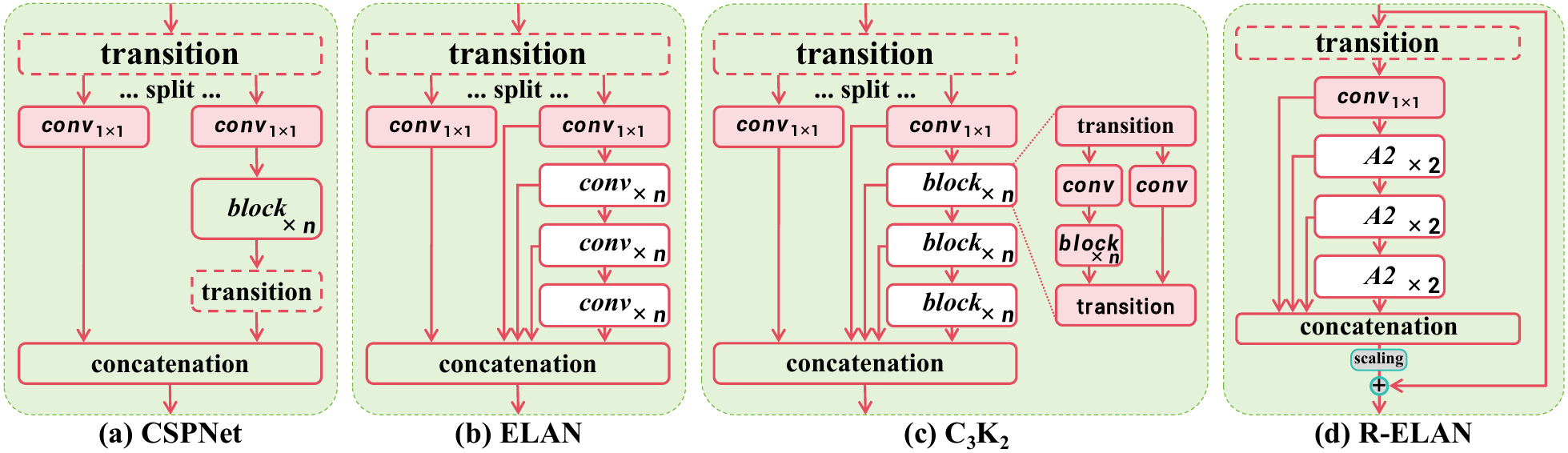}
\caption{\textbf{The architecture comparison with popular modules} including (a): CSPNet~\cite{wang2020cspnet}, (b) ELAN~\cite{wang2022designing_elan}, (c) C3K2 (a case of GELAN)~\cite{wang2024yolov9, jocher2024yolov11}, and (d) the proposed R-ELAN (residual efficient layer aggregation networks). }
\label{fig:archi_comparison}
\end{figure*}

\section{Approach}
This section introduces YOLOv12, an innovation in the YOLO framework from the perspective of network architecture with attention mechanism.

\subsection{Efficiency Analysis}
The attention mechanism, while highly effective in capturing global dependencies and facilitating tasks such as natural language processing~\cite{gpt3, devlin2018bert} and computer vision~\cite{fang2023eva, liu2021swin}, is inherently slower than convolution neural networks (CNN). Two primary factors contribute to this discrepancy in speed.

\textbf{Complexity.} First, the computational complexity of the self-attention operation scales quadratically with the input sequence length $L$. Specifically, for an input sequence with length $L$ and feature dimension $d$, the computation of the attention matrix requires $\mathcal{O}(L^2d)$ operations, since each token attends to every other token. In contrast, the complexity of convolution operations in CNNs scales linearly with respect to the spatial or temporal dimension, \textit{i.e.}, $\mathcal{O}(kLd)$, where $k$ is the kernel size and is typically much smaller than $L$. As a result, self-attention becomes computationally prohibitive, especially for large inputs such as high-resolution images or long sequences.

Moreover, another significant factor is that most attention-based vision transformers, due to their complex designs (\textit{e.g.}, window partitioning/reversing in Swin transformer~\cite{liu2021swin}) and the introduction of additional modules (\textit{e.g.}, positional encoding), gradually accumulate speed overhead, resulting in an overall slower speed compared to CNN architectures~\cite{liu2024vmamba}. In this paper, the design modules utilize simple and clean operations to implement attention, ensuring efficiency to the greatest extent.

\textbf{Computation.} Second, in the attention computation process, memory access patterns are less efficient compared to CNNs~\cite{dao2022flashattention, dao2023flashattentionv2}. Specifically, during self-attention, intermediate maps such as the attention map $(QK^T)$ and the softmax map ($L \times L$) need to be stored from high-speed GPU SRAM (the actual location of the computation) to high bandwidth GPU memory (HBM) and later retrieved during the computation, and the read and write speed of the former is more than $10$ times that of the latter, thus resulting in significant memory accessing overhead and increased wall-clock time\footnote{This problem has been solved by FlashAttention~\cite{dao2022flashattention, dao2023flashattentionv2}, which will be directly adopted during model design}. Additionally, irregular memory access patterns in attention introduce further latency compared to CNNs, which utilize structured and localized memory access. CNNs benefit from spatially constrained kernels, enabling efficient memory caching and reduced latency due to their fixed receptive fields and sliding-window operations.

These two factors, quadratic computational complexity and inefficient memory accessing, together render attention mechanisms slower than CNNs, particularly in real-time or resource-constrained scenarios. Addressing these limitations has become a critical area of research, with approaches such as sparse attention mechanisms and memory-efficient approximations (\textit{e.g.}, Linformer~\cite{wang2020linformer_attn} or Performer~\cite{choromanski2020rethinking}) aiming to mitigate the quadratic scaling.

\begin{table*}[!t]
\centering
\setlength{\tabcolsep}{0.5cm}
\caption{\textbf{Comparison with popular state-of-the-art real-time object detectors. All results are obtained using $640\times640$ inputs.} }
\selectfont
\begin{tabular}{lcccccccc}
\toprule
\textbf{\multirow{2}{*}{Method}} & \textbf{FLOPs} & \textbf{\#Param.} & \textbf{$\text{AP}^{val}_{50:95}$} &\textbf{$\text{AP}^{val}_{50}$}  &\textbf{$\text{AP}^{val}_{75}$} & \textbf{Latency}\\
& \textbf{(G)} & \textbf{(M)} & \textbf{(\%)} &\textbf{(\%)}  &\textbf{(\%)} & \textbf{(ms)}\\
\midrule
YOLOv6-3.0-N~\cite{li2023yolov6} & 11.4  & 4.7  & 37.0 &52.7 &-- & 2.69\\
Gold-YOLO-N~\cite{wang2024gold} &12.1  & 5.6  & 39.6  &55.7 &-- & 2.92 \\
YOLOv8-N~\cite{yolov8} & 8.7  & 3.2  & 37.4 &52.6 &40.5 &1.77 \\
YOLOv10-N~\cite{wang2024yolov10} &6.7  &2.3  & 38.5 &53.8 &41.7 &1.84 \\
YOLO11-N~\cite{jocher2024yolov11} &6.5  &2.6 &39.4 &55.3 &42.8 &1.5\\
\rowcolor{Gray}
\textbf{YOLOv12-N (Ours)} &\textbf{6.5}  &\textbf{2.6} &\textbf{40.6} &\textbf{56.7} &\textbf{43.8} &\textbf{1.64} \\
\midrule
YOLOv6-3.0-S~\cite{li2023yolov6} &45.3  &18.5 &44.3 &61.2 &-- &3.42 \\
Gold-YOLO-S~\cite{wang2024gold}&46.0 &21.5  &45.4 &62.5 &-- &3.82 \\
YOLOv8-S~\cite{yolov8} &28.6  &11.2 &45.0  &61.8 &48.7 &2.33 \\ 
RT-DETR-R18~\cite{zhao2024rtdetrs} &60.0 &20.0  &46.5 &63.8 &-- &4.58 \\
RT-DETRv2-R18~\cite{lv2024rtdetrv2} &60.0 &20.0  &47.9 &64.9 &-- &4.58 \\
YOLOv9-S~\cite{wang2024yolov9} &26.4 &7.1  &46.8  &63.4 &50.7 &-- \\
YOLOv10-S~\cite{wang2024yolov10} &21.6   &7.2 &46.3 &63.0 &50.4 &2.49 \\
YOLO11-S~\cite{jocher2024yolov11} &21.5 &9.4 &46.9 &63.9 &50.6 &2.5\\
\rowcolor{Gray}
\textbf{YOLOv12-S (Ours)} &\textbf{21.4}  &\textbf{9.3} &\textbf{48.0} &\textbf{65.0}  &\textbf{51.8}  &\textbf{2.61} \\
\midrule
YOLOv6-3.0-M~\cite{li2023yolov6} &85.8  &34.9  &49.1 &66.1 &-- &5.63\\
Gold-YOLO-M~\cite{wang2024gold} &87.5  &41.3 &49.8 &67.0 &-- &6.38\\
YOLOv8-M~\cite{yolov8} &78.9  &25.9  &50.3 &67.2 &54.7 &5.09\\
RT-DETR-R34~\cite{zhao2024rtdetrs} &100.0 &36.0   &48.9  &66.8 &-- &6.32\\
RT-DETRv2-R34~\cite{lv2024rtdetrv2} &100.0 &36.0  &49.9  &67.5 &-- &6.32\\
YOLOv9-M~\cite{wang2024yolov9} &76.3 &20.0  &51.4 &68.1 &56.1 &--\\
YOLOv10-M~\cite{wang2024yolov10} &59.1 &15.4  &51.1  &68.1 &55.8 &4.74\\
YOLO11-M~\cite{jocher2024yolov11} &68.0 &20.1 &51.5 &68.5 &55.7 &4.7\\
\rowcolor{Gray}
\textbf{YOLOv12-M (Ours)} &\textbf{67.5}  &\textbf{20.2} &\textbf{52.5} &\textbf{69.6}  &\textbf{57.1}  &\textbf{4.86} \\
\midrule
YOLOv6-3.0-L~\cite{li2023yolov6} &150.7  &59.6  &51.8 &69.2 &-- &9.02 \\
Gold-YOLO-L~\cite{wang2024gold} &151.7 &75.1  &51.8 &68.9 &-- &10.65 \\
YOLOv8-L~\cite{yolov8} &165.2 &43.7 &53.0  &69.8 &57.7 &8.06 \\
RT-DETR-R50~\cite{zhao2024rtdetrs} &136.0 &42.0  &53.1  &71.3 &-- &6.90\\
RT-DETRv2-R50~\cite{lv2024rtdetrv2} &136.0 &42.0  &53.4  &71.6 &-- &6.90\\
YOLOv9-C~\cite{wang2024yolov9} &102.1 &25.3  &53.0  &70.2 &57.8 &-- \\ 
YOLOv10-B~\cite{wang2024yolov10} &92.0  &19.1 &52.5  &69.6 &57.2 &5.74 \\ 
YOLOv10-L~\cite{wang2024yolov10} &120.3 &24.4  &53.2 &70.1 &58.1 &7.28 \\
YOLO11-L~\cite{jocher2024yolov11} &86.9  &25.3 &53.3 &70.1 &58.2 &6.2\\
\rowcolor{Gray}
\textbf{YOLOv12-L (Ours)} &\textbf{88.9}  &\textbf{26.4} &\textbf{53.7} &\textbf{70.7} & \textbf{58.5}  &\textbf{6.77} \\
\midrule
YOLOv8-X~\cite{yolov8} &257.8 &68.2 &54.0  &71.0 &58.8 &12.83 \\
RT-DETR-R101~\cite{zhao2024rtdetrs} &259.0 &76.0  &54.3  &72.7 &-- &13.5\\
RT-DETRv2-R101~\cite{lv2024rtdetrv2} &259.0 &76.0   &54.3  &72.8 &-- &13.5\\
YOLOv10-X~\cite{wang2024yolov10} &160.4 &29.5  &54.4 &71.3 &59.3 &10.70 \\
YOLO11-X~\cite{jocher2024yolov11} &194.9  &56.9 &54.6 &71.6 &59.5 &11.3\\
\rowcolor{Gray}
\textbf{YOLOv12-X (Ours)} &\textbf{199.0}  &\textbf{59.1} &\textbf{55.2} &\textbf{72.0} &\textbf{60.2} &\textbf{11.79} \\
\midrule
\end{tabular}
\label{tab:comparison}
\end{table*}

\subsection{Area Attention}
A simple approach to reduce the computational cost of vanilla attention is to use the linear attention mechanism~\cite{shen2021linear_attn, wang2020linformer_attn}, which reduces the complexity of vanilla attention from quadratic to linear. For a visual feature $f$ with dimensions $(n, h, d)$, where $n$ is the number of tokens, $h$ is the number of heads and $d$ is the head size, linear attention reduces the complexity from $2n^2hd$ to $2nhd^2$, decreasing the computational cost since $n > d$. However, linear attention suffers from global dependency degradation~\cite{katharopoulos2020transformers}, instability~\cite{choromanski2020rethinking}, and distribution sensitivity~\cite{xiong2021nystromformer}. Furthermore, due to the low-rank bottleneck~\cite{choromanski2020performer, bhojanapalli2020low}, it offers only limited speed advantages when applied to YOLO with input resolution of $640 \times 640$.

An alternative approach to effectively reduce complexity is the local attention mechanism (\textit{e.g.}, Shift window~\cite{liu2021swin}, criss-cross attention~\cite{huang2019ccnet}, and axial attention~\cite{dong2022cswin}), as shown in Figure~\ref{fig:split}, which transforms global attention into local, thus reducing computational costs. However, partitioning the feature map into windows can introduce overhead or reduce the receptive field, impacting both speed and accuracy.
In this study, we propose the simple yet efficient area attention module. As illustrated in Figure~\ref{fig:split}, the feature map with the resolution of $(H, W)$ is divided into $l$ segments of size $(\frac{H}{l}, W)$ or $(H, \frac{W}{l})$. This eliminates explicit window partitioning, requiring only a simple reshape operation, and thus achieves faster speed.
We empirically set the default value of $l$ to $4$, reducing the receptive field to $\frac{1}{4}$ of the original, yet it still maintains a large receptive field.
With this approach, the computational cost of the attention mechanism is reduced from $2n^2hd$ to $\frac{1}{2}n^2hd$. We show that despite the complexity $n^2$, this is still efficient enough to meet the real-time requirements of the YOLO system when $n$ is fixed at $640$ (it increases $n$ if the input resolution increases). 
Interestingly, we find that this modification has only a slight impact on performance but significantly improves speed.

\subsection{Residual Efficient Layer Aggregation Networks}
Efficient layer aggregation networks (ELAN)~\cite{wang2023yolov7} are designed to improve feature aggregation. As shown in Figure~\ref{fig:archi_comparison} (b), ELAN splits the output of a transition layer (a $1\times1$ convolution), processes one split through multiple modules, then concatenates the all the outputs and applies another transition layer (a $1\times1$ convolution) to align dimensions. However, as analyzed by~\cite{wang2023yolov7}, this architecture can introduce instability. 
We argue that such a design causes gradient blocking and lacks residual connections from input to output.
Furthermore, we build the network around the attention mechanism, which presents additional optimization challenges. Empirically, L- and X-scale models either fail to converge or remain unstable, even when using Adam or AdamW optimizers.

To address this problem, we propose residual efficient layer aggregation networks (R-ELAN), Figure~\ref{fig:archi_comparison} (d). In contrast, we introduce a residual shortcut from input to output throughout the block with a scaling factor (default to $0.01$).
This design is similar to layer scaling~\cite{touvron2021going_layer_scale}, which is introduced to build a deep vision transformer. However, applying layer scaling for each area attention will not conquer the optimization challenge and introduce slowdown on latency.
This shows that the introduction of the attention mechanism is not the only reason for convergence but the ELAN architecture itself, which verifies the rationale behind our R-ELAN design.

We also design a new aggregation approach as shown in Figure~\ref{fig:archi_comparison} (d). 
The original ELAN layer processes the input of the module by first passing it through a transition layer, which then splits it into two parts. One part is further processed by subsequent blocks, and finally both parts are concatenated to produce the output. 
In contrast, our design applies a transition layer to adjust the channel dimensions and produces a single feature map. This feature map is then processed through subsequent blocks followed by the concatenation, forming a bottleneck structure. This approach not only preserves the original feature integration capability, but also reduces both computational cost and parameter / memory usage.

\subsection{Architectural Improvements}
In this section, we will introduce the overall architecture and some improvements over the vanilla attention mechanism. Some of them are not initially proposed by us.

Many attention-centric vision transformers are designed with the plain-style architectures~\cite{dosovitskiy2020vit, touvron2021deit, bao2021beit, he2022masked, fang2023eva, fang2024eva02}, while we retain the hierarchical design of the previous YOLO systems~\cite{redmon2016yolov1, redmon2017yolo9000, redmon2018yolov3, bochkovskiy2020yolov4, jocher2020yolov5, li2023yolov6, wang2023yolov7, yolov8, wang2024yolov9, wang2024yolov10, jocher2024yolov11} and will demonstrate the necessity of this.
We remove the design of stacking three blocks in the last stage of the backbone, which is present in recent versions~\cite{yolov8, wang2024yolov9, wang2024yolov10, jocher2024yolov11}. Instead, we retain only a single R-ELAN block, reducing the total number of blocks and contributing to optimization.
We inherit the first two stages of the backbone from YOLOv11~\cite{jocher2024yolov11} and do not use the proposed R-ELAN.

Additionally, we modify several default configurations in the vanilla attention mechanism to better suit the YOLO system. These modifications include adjusting the MLP ratio from $4$ to $1.2$ (or $2$ for the N- / S- / M-scale models) is used to better allocate computational resources for better performance, adopting \texttt{nn.Conv2d+BN} instead of \texttt{nn.Linear+LN} to fully exploit the efficiency of convolution operators, removing positional encoding, and introduce a large separable convolution ($7\times7$) (namely position perceiver) to help the area attention perceive position information. The effectiveness of these modifications will be validated in Section~\ref{experiments:analysis}.

\section{Experiment}
\label{sec:experiments}

\begin{table*}[!t]
\setlength{\tabcolsep}{0.2cm}
\small
\centering
\caption{\textbf{Ablation on the proposed residual efficient layer aggregation networks (R-ELAN).} Vanilla: Uses the original ELAN design; Re-Aggre.: Employs our proposed feature integration method; Resi.: Utilizes the residual block technique; Scaling: The scaling factor for the residual connection. }
\begin{tabular}{l|cccc|c|ccc}
\toprule
\textbf{Model}  &\textbf{Vanilla}  & \textbf{Re-Aggre.}   &\textbf{Resi.}   & \textbf{Scaling}  & \textbf{Convergence}  & \textbf{FLOPs (G)} & \textbf{\#Param. (M)}  & \textbf{$\text{AP}^{val}_{50:95}$ (\%)} \\
\midrule
\multirow{3}{*}{YOLOv12-N}  & \tick   &\cross  &\cross  &--  &\tick  &6.9  &2.7  &40.8    \\
                            & \cross  &\tick   &\cross  &--  &\tick  &6.5  &2.6  &40.6  \\
                            & \cross  &\cross  &\tick   &0.1 &\tick  &6.5  &2.6  &40.3   \\
\midrule
\multirow{5}{*}{YOLOv12-L}  & \tick   &\cross  &\cross  &--  &\cross &--  &--  &--   \\
                            & \cross  &\tick   &\cross  &--  &\cross &--  &--  &--   \\
                            & \cross  &\tick   &\tick   &0.1 &\tick  &88.9  &26.4  &53.3  \\
                            & \cross  &\tick   &\tick   &0.01 &\tick &88.9  &26.4  &53.7  \\
                            & \cross  &\cross  &\tick   &0.01 &\tick &94.3  &27.8  &53.8  \\
\midrule
\multirow{5}{*}{YOLOv12-X}  & \tick   &\cross  &\cross  &--  &\cross &--  &--  &--   \\
                            & \cross  &\tick   &\cross  &--  &\cross &--  &--  &--   \\
                            & \cross  &\tick   &\tick   &0.1 &\cross &--  &--  &--  \\
                            & \cross  &\tick   &\tick   &0.01 &\tick &199.0 &59.1 &55.2  \\
                            & \cross  &\cross  &\tick   &0.01 &\tick &211.3 &62.3 &55.3  \\
\bottomrule
\end{tabular}
\label{tab:elan_ablation}
\end{table*}

This section is divided into four parts: experimental setup, a systematic comparison with popular methods, ablation studies to validate our approach, and an analysis with visualizations to further explore YOLOv12.

\subsection{Experimental Setup}
\label{experiments:details}
We validate the proposed method on the MSCOCO 2017 dataset~\cite{lin2014microsoft}. The YOLOv12 family includes $5$ variants: YOLOv12-N, YOLOv12-S, YOLOv12-M, YOLOv12-L, and YOLOv12-X. All models are trained for $600$ epochs using the SGD optimizer with an initial learning rate of $0.01$, which remains the same as YOLOv11~\cite{jocher2024yolov11}. We adopt a linear learning rate decay schedule and perform a linear warm-up for the first $3$ epochs. Following the approach in~\cite{wang2024yolov10, zhao2024rtdetrs}, the latencies of all models are tested on a T4 GPU with TensorRT FP16. 

\textbf{Baseline.}
We choose the previous version of YOLOv11~\cite{jocher2024yolov11} as our baseline. The model scaling strategy is also consistent with it. We use several of its proposed C3K2 blocks (that is a special case of GELAN~\cite{wang2024yolov9}). We do not use any more tricks beyond YOLOv11~\cite{jocher2024yolov11}.

\begin{table}[!t]
\setlength{\tabcolsep}{0.22cm}
\small
\centering
\caption{\textbf{Ablation on the proposed area attention}. With area attention (\tick), YOLOv12-N/S/X models run significantly faster on both GPU (CUDA) and CPU. CUDA results are measured on RTX 3080/A5000. Inference latency: milliseconds (ms) for FP32 and FP16 precision. (All results are obtained without using FlashAttention~\cite{dao2022flashattention, dao2023flashattentionv2}.)}
\begin{tabular}{l|c|cc|cc}
\toprule
\multirow{2}{*}{\textbf{Model}}  & \multirow{2}{*}{\textbf{\tabincell{l}{Area\\ Attention}}}  &\multicolumn{2}{c|}{\textbf{CUDA}} &\multirow{2}{*}{\textbf{CPU}}\\
& & \textbf{FP32} & \textbf{FP16} & \\
\midrule
\multirow{2}{*}{YOLOv12-N}     &\cross   &2.7/2.5   &1.5/1.5   &62.9    \\
                               &\tick    &2.0/2.0   &1.3/1.3   &31.4    \\
\midrule
\multirow{2}{*}{YOLOv12-S}     &\cross   &5.1/4.4   &2.5/2.2   &130.0   \\
                               &\tick    &3.5/3.1   &1.7/1.7   &78.4    \\
\midrule
\multirow{2}{*}{YOLOv12-X}     &\cross   &26.4/21.9  &11.1/10.4  &804.2   \\
                               &\tick    &18.2/14.3  &7.1/6.7   &512.5   \\
\bottomrule
\end{tabular}
\label{tab:area_attention}
\end{table}

\begin{table}[!htb]
\setlength{\tabcolsep}{0.14cm}
\small
\centering
\caption{\textbf{Comparative analysis of inference speed across different GPUs (RTX 3080, RTX A5000, and RTX A6000)}. Inference latency: milliseconds (ms) for FP32 and FP16 precision.}
\begin{tabular}{l|cc|cccccc}
\toprule
\multirow{2}{*}{\textbf{Model}} &\multirow{2}{*}{\textbf{Scale}} &\textbf{FLOPs}  & \multirow{2}{*}{\textbf{RTX 3080}} & \multirow{2}{*}{\textbf{A5000}} & \multirow{2}{*}{\textbf{A6000}} \\
& & (G)  & &  &  \\
\midrule
\multirow{5}{*}{\tabincell{l}{YOLOv9\\~\cite{wang2024yolov9}}}
                             &T  &8.2    & 2.4/1.5    & 2.4/1.6  & 2.3/1.7   \\
                             &S  &26.4   & 3.7/1.9    & 3.4/2.0  & 3.5/1.9   \\
                             &M  &76.3   & 6.5/2.8    & 5.5/2.6  & 5.2/2.6   \\
                             &C  &102.1  & 8.0/2.9    & 6.4/2.7  & 6.0/2.7   \\
                             &E  &189.0  & 17.2/6.7   & 14.2/6.3  & 13.1/5.9   \\
\midrule
\multirow{4}{*}{\tabincell{l}{YOLOv10\\~\cite{wang2024yolov10}}}
                             &N  &6.7   & 1.6/1.0   & 1.6/1.0  & 1.6/1.0   \\
                             &S  &21.6  & 2.8/1.4   & 2.4/1.4  & 2.4/1.3   \\
                             &M  &59.1  & 5.7/2.5   & 4.5/2.4  & 4.2/2.2   \\
                             &B  &92.0  & 6.8/2.9   & 5.5/2.6  & 5.2/2.8   \\
\midrule
\multirow{5}{*}{\tabincell{l}{YOLOv11\\~\cite{jocher2024yolov11}}}      
                             &N  &6.5    & 1.6/1.0    & 1.6/1.0  & 1.5/0.9   \\
                             &S  &21.5   & 2.8/1.3    & 2.4/1.4  & 2.4/1.3   \\
                             &M  &68.0   & 5.6/2.3    & 4.5/2.2  & 4.4/2.1   \\
                             &L  &86.9   & 7.4/3.0    & 5.9/2.7  & 5.8/2.7   \\
                             &X  &194.9  & 15.2/5.3   & 10.7/4.7  & 9.1/4.0   \\
\midrule
\multirow{5}{*}{YOLOv12}     &N  &6.5    & 1.7/1.1   & 1.7/1.0  & 1.7/1.1   \\
                             &S  &21.4   & 2.9/1.5   & 2.5/1.5  & 2.5/1.4   \\
                             &M  &67.5   & 5.8/1.5   & 4.6/2.4  & 4.4/2.2   \\
                             &L  &88.9   & 7.9/3.3   & 6.2/3.1  & 6.0/3.0   \\
                             &X  &199.0  & 15.6/5.6  & 11.0/5.2  & 9.5/4.4   \\
\bottomrule
\end{tabular}
\label{tab:speed_different_gpus}
\end{table}

\subsection{Comparison with State-of-the-arts}
\label{experiments:comparison}

We present a performance comparison between YOLOv12 and other popular real-time detectors in Table~\ref{tab:comparison}.

\textbf{For N-scale models}, YOLOv12-N outperforms YOLOv6-3.0-N~\cite{li2023yolov6}, YOLOv8-N~\cite{wang2024yolov9}, YOLOv10-N~\cite{wang2024yolov10}, and YOLOv11~\cite{jocher2024yolov11} by $3.6\%$, $3.3\%$, $2.1\%$, and $1.2\%$ in mAP, respectively, while maintaining similar or even fewer computations and parameters, and achieving the fast latency speed of $1.64$ ms/image. 

\textbf{For S-scale models}, YOLOv12-S, with $21.4$G FLOPs and $9.3$M parameters, achieves $48.0$ mAP with $2.61$ ms/image latency. It outperforms YOLOv8-S~\cite{yolov8}, YOLOv9-S~\cite{wang2024yolov9}, YOLOv10-S~\cite{wang2024yolov10}, and YOLOv11-S~\cite{jocher2024yolov11} by $3.0\%$, $1.2\%$, $1.7\%$, and $1.1\%$, respectively, while maintaining similar or fewer computations. Compared to the end-to-end detectors RT-DETR-R18~\cite{zhao2024rtdetrs} / RT-DETRv2-R18~\cite{lv2024rtdetrv2}, YOLOv12-S achieves beatable performance but with much better inference speed and less computational cost and fewer parameters.

\textbf{For M-scale models}, YOLOv12-M, with $67.5$G FLOPs and $20.2$M parameters, achieves $52.5$ mAP performance and $4.86$ ms/image speed. Compared to Gold-YOLO-M~\cite{wang2024gold}, YOLOv8-M~\cite{yolov8}, YOLOv9-M~\cite{wang2024yolov9}, YOLOv10~\cite{wang2024yolov10}, YOLOv11~\cite{jocher2024yolov11}, and RT-DETR-R34~\cite{zhao2024rtdetrs} / RT-DETRv2-R34~\cite{lv2024rt_detrv2}, YOLOv12-S enjoys superiority.

\textbf{For L-scale models}, YOLOv12-L even surpasses YOLOv10-L~\cite{wang2024yolov10} with $31.4$G fewer FLOPs. YOLOv12-L beats YOLOv11~\cite{jocher2024yolov11} by $0.4\%$ mAP with comparable FLOPs and parameters. YOLOv12-L also outperforms RT-DERT-R50~\cite{zhao2024rtdetrs} / RT-DERTv2-R50~\cite{lv2024rtdetrv2} with faster speed, fewer FLOPs ($34.6\%$) and fewer parameters ($37.1\%$). 

\textbf{For X-scale models}, YOLOv12-X significantly outperforms YOLOv10-X~\cite{wang2024yolov10} / YOLOv11-X~\cite{jocher2024yolov11} by $0.8\%$ and $0.6\%$, respectively, with comparable speed, FLOPs, and parameters. YOLOv12-X again beats RT-DETR-R101~\cite{zhao2024rtdetrs} / RT-DETRv2-R101~\cite{lv2024rt_detrv2} with faster speed, fewer FLOPs ($23.4\%$) and fewer parameters ($22.2\%$). 

In particular, if the L- / X-scale models are evaluated using FP32 precision (which requires saving the model separately in FP32 format), YOLOv12 will achieve an improvement of $\sim$ $0.2\%$ mAP. This means that YOLOv12-L/X will report $33.9\%$/$55.4\%$ mAP.

\newcommand{\subtablewidth}{0.24\textwidth}
\begin{table*}[!htb]
\centering
\setlength{\tabcolsep}{0.06cm}
\caption{\textbf{Diagnostic studies.} We only show the factor(s) to be diagnosed in each subtable to save space. The default parameters are (unless otherwise specified): training for $600$ epochs from scratch, using YOLOv12-N model.}
\begin{subtable}[t]{\subtablewidth}
\centering
\begin{tabular}[t]{l|c|c}
\textbf{Method} & \textbf{$\text{AP}^{val}_{50:95}$ } & \textbf{Latency} \\
\midrule
Linear$_{+\text{LN}}$ & 40.5  & 1.68 \\
Linear$_{+\text{BN}}$ & 39.5  & 1.70 \\
\rowcolor{Gray}
Conv$_{+\text{BN}}$ &  40.6 & 1.64 \\
Conv$_{+\text{LN}}$ &  40.3 & 1.66 \\
\end{tabular}
\caption{Attention Implementation}
\label{tab:diagnosis_attention_implementation}
\end{subtable}
\hfill
\begin{subtable}[t]{\subtablewidth}
\centering
\begin{tabular}[t]{l|c|c}
\textbf{Method} & \textbf{$\text{AP}^{val}_{50:95}$ } & \textbf{Latency} \\
\midrule
N/A    & 38.3   & 1.60 \\
\sout{S$_1$} & 40.1  & 1.63 \\
\sout{S$_4$} & 39.8  & 1.71 \\
\midrule
\rowcolor{Gray}
Ours   & 40.6  & 1.64 \\
\end{tabular}
\caption{Hierarchical Design}
\label{tab:diagnosis_hierarchical_design}
\end{subtable}
\hfill
\begin{subtable}[t]{\subtablewidth}
\centering
\begin{tabular}[t]{l|c|c}
\textbf{Epochs} & \textbf{$\text{AP}^{val}$ (N) } & \textbf{$\text{AP}^{val}$ (S)}\\
\midrule
300    & 39.5  & 47.0 \\
500    & 40.3  & 47.8 \\
\rowcolor{Gray}
600    & 40.6  & 48.0 \\
800    & 40.4  & 47.7 \\
\end{tabular}
\caption{Training Epoch}
\label{tab:diagnosis_training_epoch}
\end{subtable}
\hfill\hfill\hfill\hfill\hfill\hfill\hfill\hfill\hfill\hfill\hfill
\begin{subtable}[t]{\subtablewidth}
\centering
\begin{tabular}[t]{c|c|c}
\textbf{kernel} & \textbf{$\text{AP}^{val}_{50:95}$ } & \textbf{Latency}\\
\midrule
$3\times3$  & 40.4 & 1.60 \\
$5\times5$  & 40.4 & 1.61 \\
\rowcolor{Gray}
$7\times7$  & 40.6 & 1.64 \\
$9\times9$  & 40.7 & 1.79 \\
\end{tabular}
\caption{Position Perceiver}
\label{tab:diagnosis_perceiver}
\end{subtable}
\hfill
\\
\begin{subtable}[t]{\subtablewidth}
\centering
\begin{tabular}[t]{l|c|c}
\textbf{Pos.} & \textbf{$\text{AP}^{val}_{50:95}$ } & \textbf{Latency} \\
\midrule
RPE & 40.3  & 1.76 \\
APE & 40.5  & 1.69 \\
\midrule
\rowcolor{Gray}
N/A & 40.6  & 1.64 \\
\end{tabular}
\caption{Position Embedding}
\label{tab:diagnosis_position_embeddings}
\end{subtable}
\hfill
\vspace{0.2cm}
\begin{subtable}[t]{\subtablewidth}
\centering
\begin{tabular}[t]{c|c|c}
\textbf{Area}  & \textbf{$\text{AP}^{val}_{50:95}$ } & \textbf{Latency} \\
\midrule
\cross  &40.8 & 1.70 \\
\rowcolor{Gray}
\tick   &40.6 & 1.64 \\
\end{tabular}
\caption{Area Attention}
\label{tab:diagnosis_area_attention}
\end{subtable}
\hfill
\begin{subtable}[t]{\subtablewidth}
\centering
\begin{tabular}[t]{c|c|c}
\textbf{Ratio (L)} & \textbf{$\text{AP}^{val}_{50:95}$ } & \textbf{Latency}\\
\midrule
\rowcolor{Gray}
1.2 & 53.8 & 6.77 \\
2.0 & 53.6 & 6.75 \\
4.0 & 53.1 & 6.68 \\
\end{tabular}
\caption{MLP Ratio}
\label{tab:diagnosis_mlp_ratio}
\end{subtable}
\hfill
\begin{subtable}[t]{\subtablewidth}
\centering
\begin{tabular}[t]{c|c|c}
\textbf{FA} & \textbf{Latency (N)} & \textbf{Latency (S)}\\
\midrule
\cross  & 1.92 & 3.02 \\
\rowcolor{Gray}
\tick   & 1.64 & 2.61 \\
\end{tabular}
\caption{FlashAttention}
\label{tab:diagnosis_flashattention}
\end{subtable}
\label{tab:diagnosis}
\end{table*}

\subsection{Ablation Studies}
\label{experiments:ablation}

\noindent$\bullet$\hspace{0.2cm}\textbf{R-ELAN.}
Table~\ref{tab:elan_ablation} evaluates the effectiveness of the proposed residual efficient layer networks (R-ELAN) using YOLOv12-N/L/X models. The results reveal two key findings:  
\textbf{(i)} For small models like YOLOv12-N, residual connections do not impact convergence but degrade performance. In contrast, for larger models (YOLOv12-L/X), they are essential for stable training. In particular, YOLOv12-X requires a minimal scaling factor ($0.01$) to ensure convergence.  
\textbf{(ii)} The proposed feature integration method effectively reduces the complexity of the model in terms of FLOPs and parameters while maintaining comparable performance with only a marginal decrease.

\noindent$\bullet$\hspace{0.2cm}\textbf{Area Attention.}
We conduct ablation experiments to validate the effectiveness of area attention, with results presented in Table~\ref{tab:area_attention}. Evaluations are performed on YOLOv12-N/S/X models, measuring the inference speed on both the GPU (CUDA) and CPU. CUDA results are obtained using RTX 3080 and A5000, while CPU performance is measured on an Intel Core i7-10700K @ 3.80GHz. The results demonstrate a significant speedup with area attention (\tick). For example, with FP32 on RTX 3080, YOLOv12-N achieves a $0.7$ms reduction in inference time. This performance gain is consistently observed across different models and hardware configurations.
We do not use FlashAttention~\cite{dao2022flashattention, dao2023flashattentionv2} in this experiment because it would significantly reduce the speed difference.

\begin{figure*}[!htb]
\centering
\includegraphics[width=0.99\linewidth]{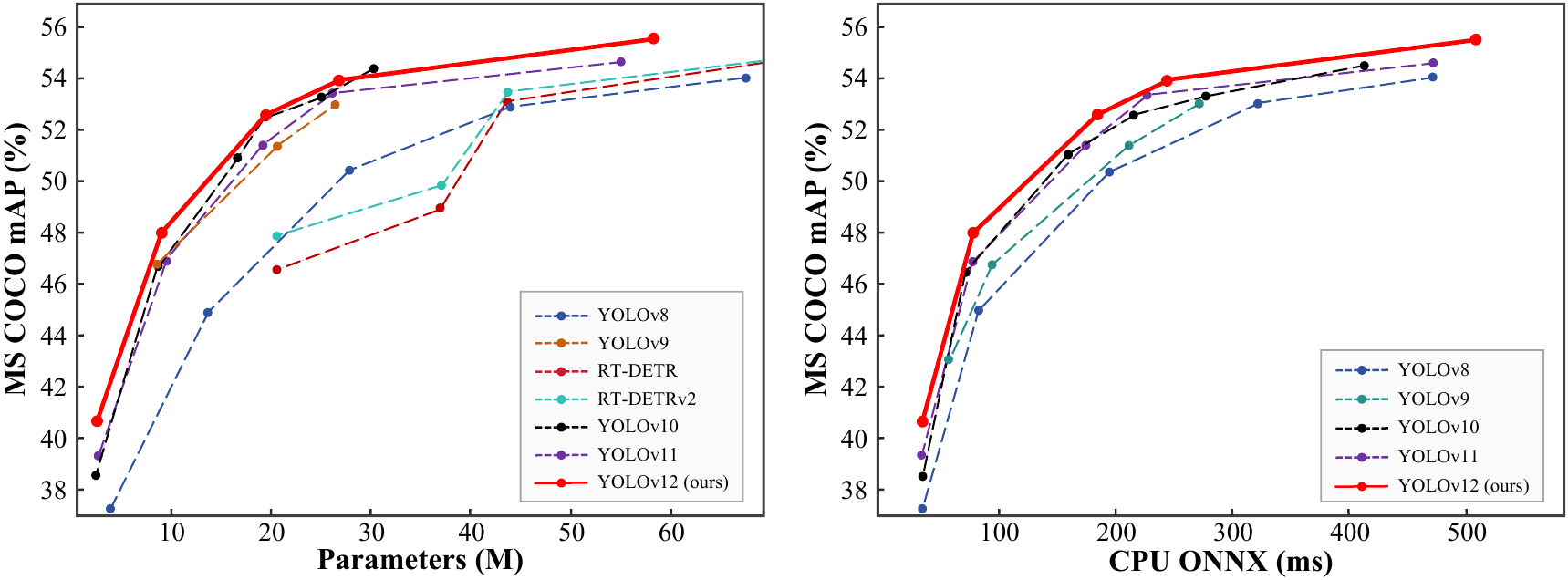}
\caption{\textbf{Comparison with popular methods in terms of accuracy-parameters (left) and accuracy-latency trade-off on CPU (right).}}
\label{fig:cpu_speed_comparison}
\end{figure*}

\subsection{Speed Comparison}
\label{experiments:speed}

Table~\ref{tab:speed_different_gpus} presents a comparative analysis of inference speed across different GPUs, evaluating YOLOv9~\cite{wang2024yolov9}, YOLOv10~\cite{wang2024yolov10}, YOLOv11~\cite{jocher2024yolov11}, and our YOLOv12 on RTX 3080, RTX A5000, and RTX A6000 with FP32 and FP16 precision. To ensure consistency, all results are obtained on the same hardware and YOLOv9~\cite{wang2024yolov9} and YOLOv10~\cite{wang2024yolov10} are evaluated using the integrated codebase of ultralytics~\cite{jocher2024yolov11}.
The results indicate that YOLOv12 achieves a significantly higher inference speed than YOLOv9~\cite{wang2024yolov9} while remaining on par with YOLOv10~\cite{wang2024yolov10} and YOLOv11~\cite{jocher2024yolov11}. For example, on RTX 3080, YOLOv9 reports $2.4$ ms (FP32) and $1.5$ ms (FP16), while YOLOv12-N achieves $1.7$ ms (FP32) and $1.1$ ms (FP16). Similar trends hold across other configurations.

Figure~\ref{fig:cpu_speed_comparison} presents additional comparisons. The left subfigure illustrates the accuracy-parameter trade-off comparison with popular methods, where YOLOv12 establishes a dominant boundary beyond the counterparts, surpassing even YOLOv10, A YOLO version characterized by significantly fewer parameters, showcasing the efficacy of YOLOv12.
We compare the inference latency of YOLOv12 with previous YOLO versions on a CPU in the right subfigure (All results were measured on an Intel Core i7-10700K @ 3.80GHz). As shown, YOLOv12 surpasses other competitors with a more advantageous boundary, highlighting its efficiency across diverse hardware platforms.

\subsection{Diagnosis \& Visualization}
\label{experiments:analysis}

We diagnose the YOLOv12 designs in Tables~\ref{tab:diagnosis_attention_implementation} to~\ref{tab:diagnosis_flashattention}. Unless otherwise specified, we perform these diagnostics on YOLOv12-N, with a default training of $600$ epochs from scratch.

\begin{figure*}[!t]
\centering
\includegraphics[width=1\linewidth]{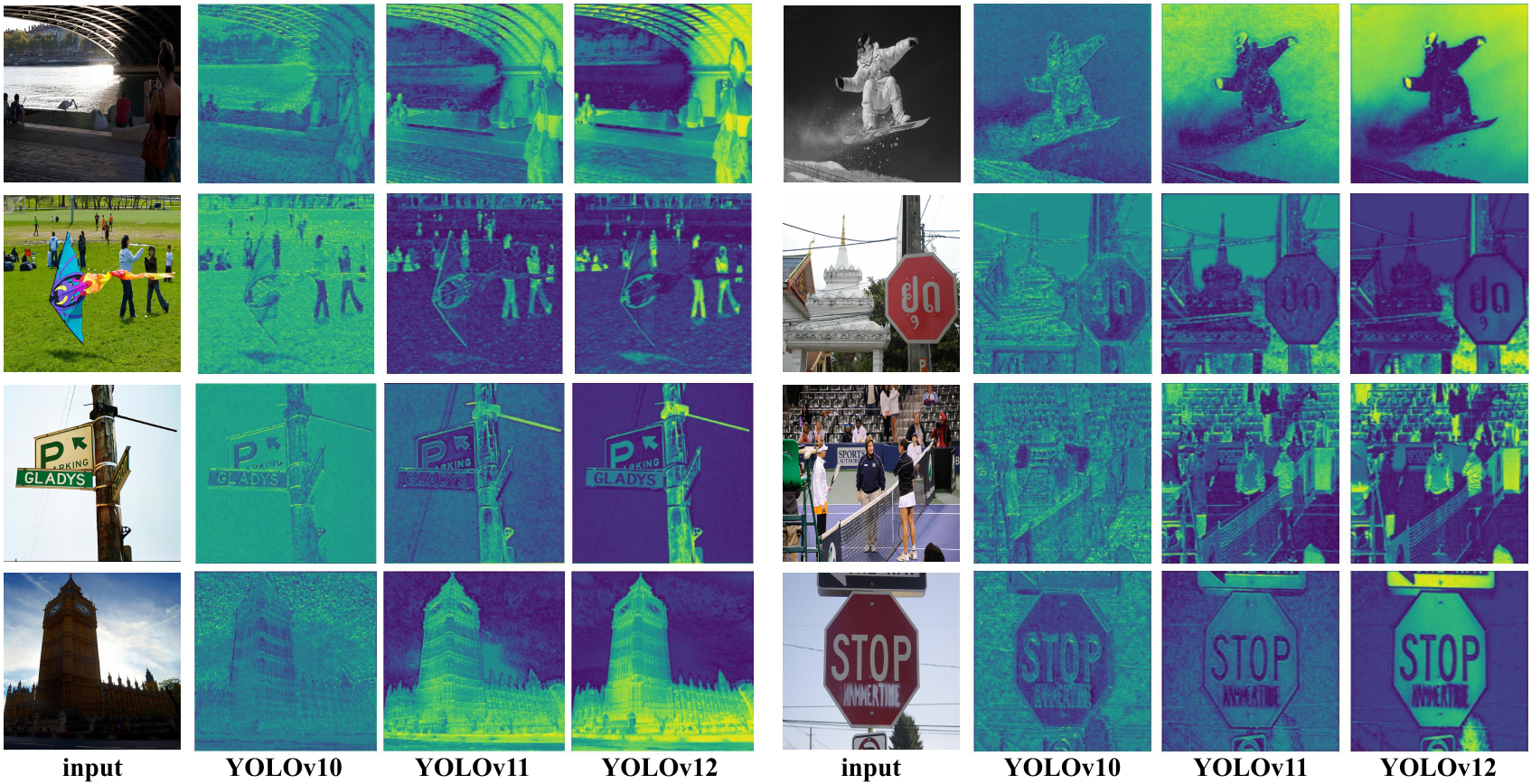}
\caption{\textbf{Comparison of heat maps between YOLOv10~\cite{wang2024yolov10}, YOLOv11~\cite{jocher2024yolov11}, and the proposed YOLOv12.} Compared to the advanced YOLOv10 and YOLOv11, YOLOv12 demonstrates a clearer perception of objects in the image. All the results are obtained using the X scale models. \textit{Zoom in to compare the details.}}
\label{fig:heat_map}
\end{figure*}

\noindent$\bullet$\hspace{0.2cm}\textbf{Attention Implementation: Table~\ref{tab:diagnosis_attention_implementation}.}\quad
We examine two approaches to implementing attention. The convolution-based approach is faster than the linear-based approach due to the computational efficiency of convolution. Additionally, we explore two normalization methods (layer normalization (LN) and batch normalization (BN)) and find the result: although layer normalization is commonly used in attention mechanisms, it performs worse than batch normalization when used with convolution. Notably, this has already been used in the PSA module~\cite{wang2024yolov10} and our finding is in line with its design.

\noindent$\bullet$\hspace{0.2cm}\textbf{Hierarchical Design: Table~\ref{tab:diagnosis_hierarchical_design}.}\quad
Unlike other detection systems such as Mask R-CNN~\cite{he2022masked, bao2021beit}, where the architectures of the plain vision transformers can produce strong results, YOLOv12 exhibits a different behavior. When using a plain vision transformer (N/A), the detector's performance drops significantly, achieving only $38.3\%$ mAP. A more moderate adjustment, such as omitting the first (\sout{S$_1$}) or fourth stage (\sout{S$_4$}), while maintaining similar FLOPs by adjusting the feature dimensions, results in a slight performance degradation of $0.5\%$ mAP and $0.8\%$ mAP, respectively. Consistent with previous YOLO models, the hierarchical design remains the most effective, yielding the best performance in YOLOv12.

\noindent$\bullet$\hspace{0.2cm}\textbf{Training Epochs: Table~\ref{tab:diagnosis_training_epoch}.}\quad
We examine how varying the number of training epochs impacts performance (training from scratch). Although some existing YOLO detectors achieve optimal results after roughly $500$ training epochs~\cite{yolov8, wang2024yolov9, wang2024yolov10}, YOLOv12 requires a more extended training period (about $600$ epochs) to achieve peak performance, keeping the same configuration used in YOLOv11~\cite{jocher2024yolov11}.

\noindent$\bullet$\hspace{0.2cm}\textbf{Position Perceiver: Tables~\ref{tab:diagnosis_perceiver}.}\quad
In the attention mechanism, we apply a separable convolution with a large kernel to the attention value $v$, adding its output to $v @ \text{attn}$. We refer to this component as the Position Perceiver, as the smoothing effect of convolution preserves the original positions of image pixels, it helps the attention mechanism perceive positional information (This has already been used in the PSA module~\cite{wang2024yolov10}, but we expand the convolution kernel, achieving performance improvement without affecting speed). As shown in the table, increasing the convolution kernel size improves performance but gradually reduces speed. When the kernel size reaches $9 \times 9$, the slowdown becomes significant. Therefore, we set $7 \times 7$ as the default kernel size.

\begin{table*}[!htb]
\setlength{\tabcolsep}{0.3cm}
\small
\centering
\caption{\textbf{Detailed performance of YOLOv12 on COCO.}}
\begin{tabular}{l|cccccccc}
\toprule
&\textbf{$\text{AP}^{val}_{50:95}$ (\%)}   &\textbf{$\text{AP}^{val}_{50}$}(\%) &\textbf{$\text{AP}^{val}_{75}$ (\%)}      &\textbf{$\text{AP}^{val}_{small}$ (\%)}  &\textbf{$\text{AP}^{val}_{medium}$ (\%)}  &\textbf{$\text{AP}^{val}_{large}$ (\%)}\\
\midrule
YOLOv12-N  & 40.6  & 56.7   & 43.8  & 20.2  & 45.2  & 58.4  \\
YOLOv12-S  & 48.0  & 65.0   & 51.8  & 29.8  & 53.2  & 65.6  \\
YOLOv12-M  & 52.5  & 69.6   & 57.1  & 35.7  & 58.2  & 68.8  \\
YOLOv12-L  & 53.7  & 70.7   & 58.5  & 36.9  & 59.5  & 69.9  \\
YOLOv12-X  & 55.2  & 72.0   & 60.2  & 39.6  & 60.7  & 70.9  \\
\bottomrule
\end{tabular}
\label{tab:detailed_results}
\end{table*}

\noindent$\bullet$\hspace{0.2cm}\textbf{Position Embedding: Tables~\ref{tab:diagnosis_position_embeddings}.}\quad
We examine the impact of commonly used positional embeddings in most attention-based models (RPE: relative positional embedding; APE: absolute positional encoding) on performance. Interestingly, the best-performing configuration is achieved without any positional embedding, which brings cleaner architecture and faster inference latency.

\noindent$\bullet$\hspace{0.2cm}\textbf{Area Attention: Tables~\ref{tab:diagnosis_area_attention}.}\quad
In this table, we use the FlashAttention technique by default. This causes that while the area attention mechanism increases computational complexity (leading to performance gains), the resulting slowdown remains minimal. For further validation of area attention effectiveness, refer to Table~\ref{tab:area_attention}.

\noindent$\bullet$\hspace{0.2cm}\textbf{MLP Ratio: Tables~\ref{tab:diagnosis_mlp_ratio}.}\quad
In traditional vision transformers, the MLP ratio within the attention module is generally set to $4.0$. However, we observe different behavior with YOLOv12. In the table, varying the MLP ratio impacts the model size, so we adjust the feature dimensions to maintain overall model consistency. In particular, YOLOv12 achieves better performance with an MLP ratio of $1.2$, diverging from conventional practices. This adjustment shifts the computational load more toward the attention mechanism, highlighting the importance of area attention.

\noindent$\bullet$\hspace{0.2cm}\textbf{FlashAttention: Tables~\ref{tab:diagnosis_flashattention}.}\quad
This table validates the role of FlashAttention in YOLOv12. It shows that FlashAttention accelerates YOLOv12-N by approximately $0.3$ms and YOLOv12-S by around $0.4$ms without other costs.

\noindent\textbf{Visualization: Heat Map Comparison.}
Figure~\ref{fig:heat_map} compares the heat maps of YOLOv12 with state-of-the-art YOLOv10~\cite{wang2024yolov10} and YOLOv11~\cite{jocher2024yolov11}. 
These heat maps, extracted from the third stage of the backbones of X-scale models, highlight the regions activated by the model, reflecting its object perception capability.
As illustrated, compared to YOLOv10 and YOLOv11, YOLOv12 produces clearer object contours and more precise foreground activation, indicating improved perception.
Our explanation is that this improvement comes from the area attention mechanism, which has a larger receptive field than convolutional networks and is therefore considered better at capturing the overall context, leading to more precise foreground activation. We believe that this characteristic gives YOLOv12 a performance advantage.

\begin{table}[h]
    \centering
    \setlength{\tabcolsep}{0.18cm}
    \caption{\textbf{Hyperparameters for fine-tuning the YOLOv12 family on COCO~\cite{lin2014microsoft}.}}
    \begin{tabular}{l|c}
    \toprule
    \textbf{Hyperparameters}                   & \textbf{YOLOv12-N/S/M/L/X} \\ 
    \midrule
    \multicolumn{2}{l}{\textbf{\textit{Training Configuration}}} \\
    Epochs                                    & $600$ \\
    Optimizer                                 & SGD \\
    Momentum                                  & $0.937$ \\
    Batch size                                & $32\times8$ \\
    Weight decay                              & $5 \times 10^{-4}$ \\
    Warm-up epochs                            & $3$ \\
    Warm-up momentum                          & $0.8$ \\
    Warm-up bias learning rate                & $0.0$ \\
    Initial learning rate                     & $10^{-2}$ \\
    Final learning rate                       & $10^{-4}$ \\
    Learning rate schedule                    & Linear decay \\
    \midrule
    \multicolumn{2}{l}{\textbf{\textit{Loss Parameters}}} \\
    Box loss gain                             & $7.5$ \\
    Class loss gain                           & $0.5$ \\
    DFL loss gain                             & $1.5$ \\
    \midrule
    \multicolumn{2}{l}{\textbf{\textit{Augmentation Parameters}}} \\
    HSV saturation augmentation               & $0.7$ \\
    HSV value augmentation                    & $0.4$ \\
    HSV hue augmentation                      & $0.015$ \\
    Translation augmentation                  & $0.1$ \\
    Scale augmentation                        & $0.5 / 0.9 / 0.9 / 0.9 / 0.9$ \\
    Mosaic augmentation                       & $1.0$ \\
    Mixup augmentation                        & $0.0 / 0.05 / 0.15 / 0.15 / 0.2$ \\
    Copy-paste augmentation                   & $0.1 / 0.15 / 0.4 / 0.5 / 0.6$ \\
    Close mosaic epochs                       & $10$ \\
    \bottomrule
    \end{tabular}
    \label{tab:finetuning_details}
\end{table}

\section{Conclusion}
This study introduces YOLOv12, which successfully adopts an attention-centric design that traditionally is considered inefficient for real-time requirements, into the YOLO framework, achieving a state-of-the-art latency-accuracy trade-off.  
To enable efficient inference, we propose a novel network that leverages area attention to reduce computational complexity and residual efficient layer aggregation networks (R-ELAN) to enhance feature aggregation. Furthermore, we refine key components of the vanilla attention mechanism to better align with YOLO’s real-time constraints while maintaining high-speed performance.  
As a result, YOLOv12 achieves state-of-the-art performance by effectively combining area attention, R-ELAN, and architectural optimizations, leading to significant improvements in both accuracy and efficiency. Comprehensive ablation studies further validate the effectiveness of these innovations.  
This study challenges the dominance of CNN-based designs in YOLO systems and advances the integration of attention mechanisms for real-time object detection, paving the way for a more efficient and powerful YOLO system.

\section{Limitations} YOLOv12 requires FlashAttention~\cite{dao2022flashattention, dao2023flashattentionv2}, which currently supports Turing, Ampere, Ada Lovelace, or Hopper GPUs (\textit{e.g.}, T4, Quadro RTX series, RTX20 series, RTX30 series, RTX40 series, RTX A5000/6000, A30/40, A100, H100, \textit{etc.}).

\section{More Details}

\noindent\textbf{Fine-tuning Details.} 
By default, all YOLOv12 models are trained using the SGD optimizer for $600$ epochs. Following previous works~\cite{wang2023yolov7, yolov8, wang2024yolov9, wang2024yolov10}, the SGD momentum and weight decay are set to $0.937$ and $5 \times 10^{-4}$, respectively. The initial learning rate is set to $1 \times 10^{-2}$ and decays linearly to $1 \times 10^{-4}$ throughout the training process. Data augmentations, including Mosaic~\cite{bochkovskiy2020yolov4, wang2023yolov7}, Mixup~\cite{zhu2020deformable_detr}, and copy-paste augmentation~\cite{zhang2017mixup}, are applied to enhance training.
Following YOLOv11~\cite{jocher2024yolov11}, we adopt the Albumentations library~\cite{buslaev2020albumentations}.
Detailed hyperparameters are presented in Table~\ref{tab:finetuning_details}. All models are trained on $8\times$ NVIDIA A6000 GPUs. Following established conventions~\cite{yolov8, wang2024yolov9, wang2024yolov10, jocher2024yolov11}, we report the standard mean average precision (mAP) on different object scales and IoU thresholds. In addition, we report the average latency in all images.
We recommend reviewing more details at the official code: \url{https://github.com/sunsmarterjie/yolov12}.

\noindent\textbf{Result Details.} 
We report more details of the results in Table~\ref{tab:detailed_results} including $\text{AP}^{val}_{50:95}$, $\text{AP}^{val}_{50}$, $\text{AP}^{val}_{75}$,    $\text{AP}^{val}_{small}$, $\text{AP}^{val}_{medium}$, $\text{AP}^{val}_{large}$.

{
    \small
    \bibliographystyle{ieeenat_fullname}
    \bibliography{main}
}

\end{document}